\documentclass{article}
\usepackage{spconf,amsmath,graphicx}
\usepackage[ruled]{algorithm2e}
\usepackage{amssymb}
\usepackage[inline]{enumitem}
\usepackage[table,x11names]{xcolor}
\usepackage[bookmarks=false]{hyperref}

\definecolor{yellow}{rgb}{1,1,0}

\title{Unsupervised Clustering of Time Series Signals using Neuromorphic Energy-Efficient Temporal Neural Networks }
\name{Shreyas Chaudhari \qquad Harideep Nair \qquad Jos\'e M.F. Moura \qquad John Paul Shen} 
\address{Carnegie Mellon University - Electrical and Computer Engineering Department} 
\begin{document}
%
\maketitle
\begin{abstract}
Unsupervised time series clustering is a challenging problem with diverse industrial applications such as anomaly detection, bio-wearables, etc. These applications typically involve small, low-power devices on the edge that collect and process real-time sensory signals. State-of-the-art time-series clustering methods perform some form of loss minimization that is extremely computationally intensive from the perspective of edge devices. In this work, we propose a neuromorphic approach to unsupervised time series clustering based on Temporal Neural Networks that is capable of ultra low-power, continuous online learning.
We demonstrate its clustering performance on a subset of UCR Time Series Archive datasets. Our results show that the proposed approach either outperforms or performs similarly to most of the existing algorithms while being far more amenable for efficient hardware implementation. Our hardware assessment analysis shows that in 7 nm CMOS the proposed architecture, on average, consumes only about 0.005 mm\textsuperscript{2} die area and 22 $\mu$W power and can process each signal with about 5 ns latency.
\end{abstract}
\section{Introduction}
\label{sec:intro}
Supervised deep neural networks have revolutionized the field of Machine Learning and achieve state-of-the-art results on many applications such as Computer Vision, Natural Language Processing, etc. However, they require 1) extensive amounts of labeled data, 2) large power-hungry compute resources, and 3) separate training and testing phases leading to poor generalization across unseen data patterns. These constraints become particularly pertinent for industrial \textit{edge} applications involving time-series signals such as anomaly detection, bio-wearables, smart city sensors, etc. Unsupervised time-series clustering, which groups signals according to similar features, overcomes the first limitation. The second and third limitations raise the need for online processing that can dynamically capture the changes in \textit{macro} patterns over time, within a tiny power budget.

Several techniques for time-series clustering have been studied in the literature. The work in \cite{Ma2019} proposes a deep learning based approach to learn feature representations for enhanced clustering performance. They propose an encoder-decoder architecture and train the model using a joint classification, reconstruction and K-means loss function. While their approach achieves state-of-the-art performance, it requires the deep network to be trained a-priori using synthetic data. Furthermore, the approach, involving multi-layer RNNs and backpropagation, is not suitable for computationally and memory constrained edge devices.
A method for learning shapelets, or discriminative segments of a time series, in an unsupervised manner has been proposed in \cite{Zhang2018}. While typical shapelet based methods exhaustively search a set of candidates to find the optimal shapelets for clustering, the USSL algorithm described in \cite{Zhang2018} builds on \cite{grabocka2014learning} to formulate the shapelet learning problem as an optimization problem that can be solved using coordinate descent. However, the authors use pseudo-labels to iteratively improve the quality of the discovered shapelets. While there are many other such works on unsupervised clustering in literature, low-power online learning algorithms have not been widely explored.

In this paper we present a low-power, online, unsupervised processor for time-series clustering based on the recently proposed Temporal Neural Networks (TNNs). TNNs are a specific class of Spiking Neural Networks (SNNs) that encode and process information \textit{temporally} via relative spike timing relationships and are defined by a rigorous space-time algebra \cite{smith2018space, SmithBook}. The space-time computing framework of TNNs resemble that of the mammalian neocortex and can be directly implemented using off-the-shelf CMOS technology \cite{nair2020direct}.
To the best of our knowledge, this is the first work to propose a TNN-based approach for time-series clustering.

\section{Background}
\label{sec:background}

\subsection{SNNs and TNNs}
\label{ssec:tnn}

\begin{figure}[t]
    \centering
    \includegraphics[width=245pt]{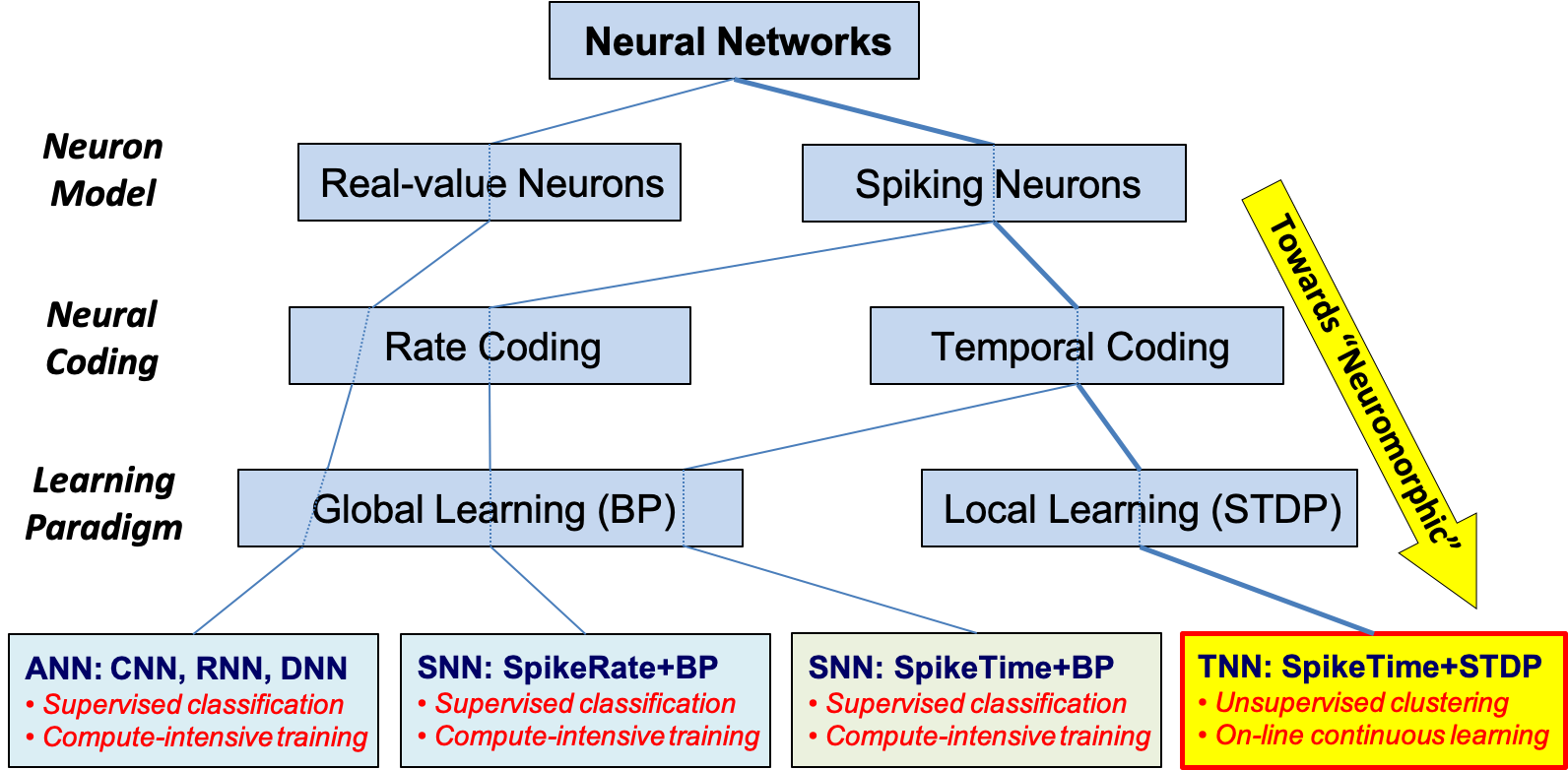}
    \caption{Neural Network Taxonomy \cite{nair2020direct}}
    \label{fig:taxonomy}
\end{figure}

Biologically inspired spiking neural networks have become increasingly popular due to their ability to perform pattern recognition tasks with ultra-low power consumption. These networks are typically composed of an encoding layer, one or more fully connected layers, and most recently, convolution layers \cite{diehl2015unsupervised, tavanaei2016bio, tavanaei2017multi, kheradpisheh2018stdp, tavanaei2019deep, mozafari2019bio}. 
Most spiking neural networks in the literature either encode the input stimulus as the spiking rate and/or use backpropagation for learning (Figure \ref{fig:taxonomy}), both of which are not biologically plausible. Temporal neural networks (TNNs) employ intensity-to-latency temporal encoding and Spike Timing Dependent Plasticity (STDP) learning, both of which are biologically plausible. The architecture of TNNs is composed of a hierarchy of building blocks, namely, multi-synapse \textit{neurons}, multi-neuron \textit{columns} and multi-column \textit{layers} and can be directly implemented in hardware using standard CMOS technology \cite{nair2020direct}. Synapses in TNN networks store weights and these weights are trained using STDP.
\subsection{Temporal Encoding}
\label{ssec:subsubhead}
Two neuron coding schemes have been primarily used in SNNs: \textit{rate coding} and \textit{temporal coding}. As implied by their name, rate-based coding schemes represent information using the frequency of spikes within a time interval and hence emit multiple spikes for a single input. Temporal coding methods, however, represent information in the relative time at which a spike occurs, rather than the spike frequency, leading to a single spike per input. Temporal codes are thus capable of accurately representing data with sparse spiking activity, and consume far less power than rate codes \cite{SmithBook}. Furthermore, it has been experimentally shown in \cite{Pan2019} that in some applications, encoding data with a population of neurons is more accurate than using a single neuron alone, i.e., it offers greater resolution for data representation. Hence, in this work, we apply both temporal and population coding mechanisms for representing time series signals as spike trains.
\section{Proposed Approach}
\begin{figure}[ht]
    \centering
    \includegraphics[width=250pt]{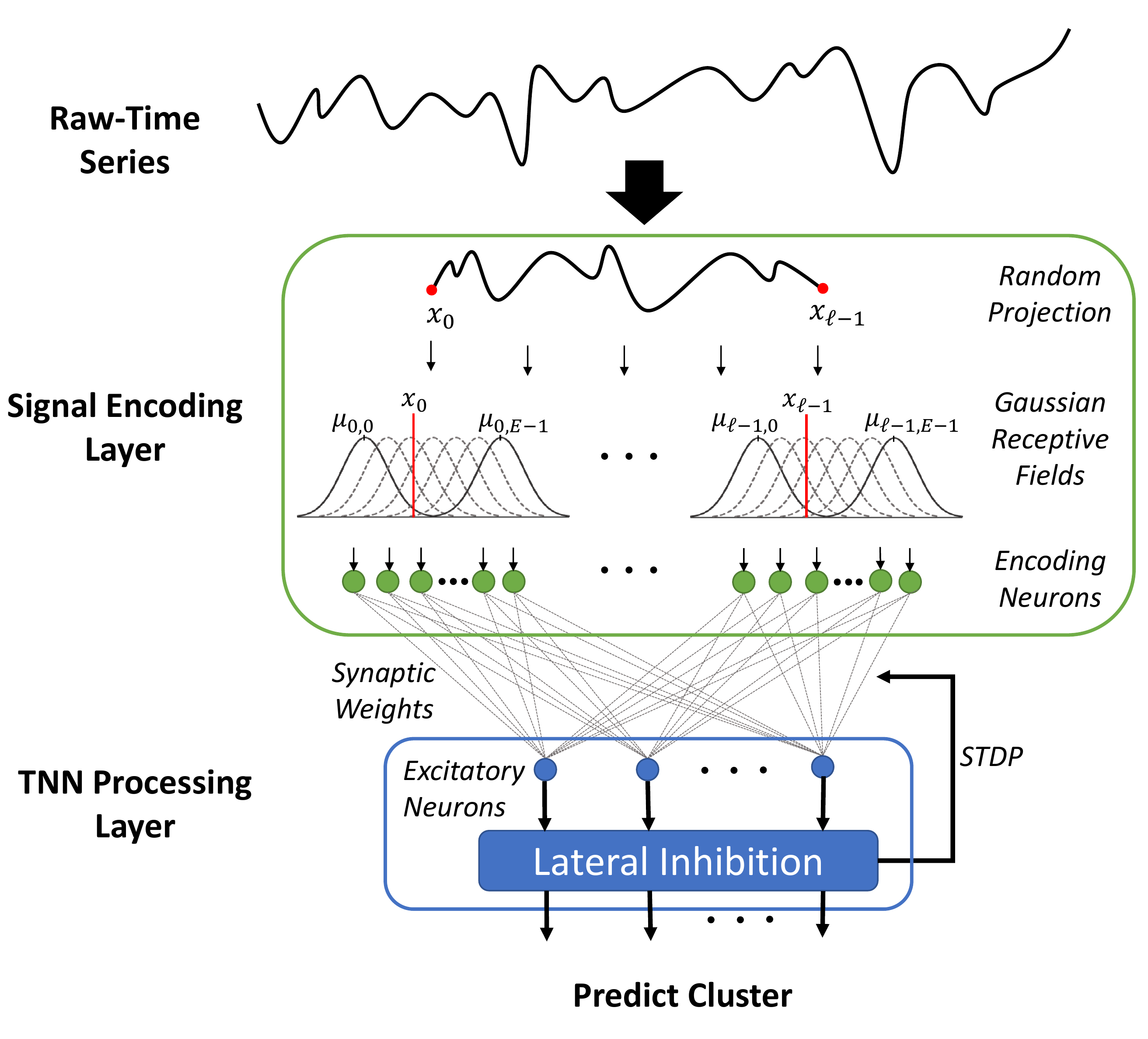}
    \caption{Block diagram of the proposed TNN architecture}
    \label{fig:arch_block_daigram}
\end{figure}

A block diagram of the proposed unsupervised time series clustering architecture is shown in Figure \ref{fig:arch_block_daigram}. The main components of our architecture are described in detail below.
\subsection{Signal Encoding Layer}
The encoding layer transforms the input time-series signal into a series of spike trains. Suppose the training dataset $\mathbf{D} \in \mathbb{R}^{N \times L}$ contains $N$ uni-variate signals, where each signal is of length $L$. We first perform dimensionality reduction on the incoming signals using sparse random projections. The Johnson-Lindenstrauss lemma \cite{johnson1984extensions} asserts that a set of points in $n$-dimensional Euclidean space can be randomly projected onto a $p$-dimensional linear subspace where $p < n$ such that the pairwise distances between points are roughly preserved. Achlioptis \cite{Achlioptas2003} showed that the transformation could be computed using a sparse matrix $\mathbf{P}$ whose $ij$-element can take one of the three values $\sqrt{3}$, $0$, $-\sqrt{3}$ with respective probabilities $\frac{1}{6}$, $\frac{2}{3}$ and $\frac{1}{6}$.
Thus we generate the reduced dataset $\widetilde{\mathbf{D}} = \mathbf{D}\mathbf{P}$ where the projection matrix $\mathbf{P}$ has dimensions $L \times \ell$ (where $\ell << L$). The sparsity of $\mathbf{P}$ enables this lower dimensionality approach to be  implemented on resource-constrained devices. It also significantly reduces the hardware implementation complexity of the TNN processing layer. 

The transformed data $\widetilde{\mathbf{D}}$ is then converted to spike trains using Gaussian receptive fields resembling the approach in \cite{Bohte2002}.
Specifically, let $\mathbf{x} = (x_0, x_1, \dots, x_{\ell-1})$ be a data sample (row) from $\widetilde{\mathbf{D}}$. A point $x_i$ is encoded by a population of $E$ encoding neurons, each of which describes a Gaussian function. The parameters for the Gaussian functions for $x_i$ are computed over all data samples (column $i$ of $\widetilde{\mathbf{D}}$). The width $\sigma_i$ and center $\mu_{ij}$ for the $i^{th}$ time step and $j^{th}$ neuron ($j = 0, 1, \dots, E-1$) are defined as:
\begin{align*}
    \sigma_i &= \frac{\gamma\left(x_{i,\;\max} - x_{i,\;\min}\right)}{E - 2}\\
    \mu_{ij} &= x_{i,\;\min} + \left(\frac{2j-3}{2}\right)\sigma_i
\end{align*}
Here, $x_{i,\;\min}$ and $x_{i,\;\max}$ represent the minimum and maximum value in the $i^{th}$ column of $\widetilde{\mathbf{D}}$ while $\gamma$ is a scaling factor. The time $t_j^{(e)}$ at which an encoding neuron $j$ emits a spike for input $x_i$ is:
\begin{align*}
    f_j = \exp{\left[-\frac{1}{2}\left(\frac{x_i - \mu_{ij}}{\sigma_i}\right)^2\right]}\\
    t_j^{(e)} = \mathrm{round} \left[T_{\max}\cdot\left(1 - f_j\right)\right]
\end{align*}
Thus the encoding layer emits a total of $E\cdot\ell$ discrete spike trains for $\ell$ inputs. Here, $T_{\max}$ is an exclusive upper bound on the spike times generated by the encoding neurons. If $t_j^{(e)} = T_{\max}$, then no spike is emitted by the neuron. Note that our architecture model is \textit{completely integer-based} and therefore highly amenable for efficient hardware implementation.

\subsection{TNN Processing Layer}
\label{ssec:subsubhead}
\subsubsection{Excitatory Neurons}
The fully connected TNN processing layer is composed of a single column of ramp-no-leak, integrate and fire neurons \cite{nair2020direct}. The number of neurons in the layer is equal to the number of clusters, while the number of synapses per neuron is equal to $E\cdot\ell$. The body (or membrane) potential of the $k^{th}$ neuron in the TNN processing layer at time $t$ is modeled as:
\begin{align*}
    v_k(t) = \sum_{j}\rho\left(t - t_j^{(e)}, w_{kj}\right)
\end{align*}
Here, $w_{kj}$ is the synaptic weight from neuron $j$ in the encoding layer to neuron $k$ in the processing layer, and $t_j^{(e)}$ represents the input spike time from encoding neuron $j$. The ramp-no-leak response function, $\rho$, is defined as:
\begin{align*}
    \rho(t, w) = \begin{cases}0 & \text{if } t < 0\\
    t & \text{if } 0\leq t < w\\
    w & \text{if } t \geq w
    \end{cases}
\end{align*}
A neuron in the processing layer emits a spike immediately after its body potential exceeds some threshold $\theta$. 
A neuron may only emit a single spike in a forward pass for any given data sample. After each forward pass (which includes lateral inhibition described below), the body potentials for all neurons in the layer are reset. 
The output spike time $t_k$ of the $k^{th}$ neuron in the processing layer is formally computed as:
\begin{align*}
    t_k = \begin{cases} t_{out} & \text{ if }\;\exists\;t_{out} \;\mathrm{s.t.}\; v_k(t_{out}) \geq \theta \text{ and}\\
    & \nexists\;t < t_{out} \; \mathrm{s.t.}\; v_k(t) \geq \theta\\
    T_{\max} & \text{otherwise}\end{cases}
\end{align*}

\subsubsection{Lateral Inhibition}
\label{ssec:lateral_inhibition}
Lateral inhibition is a biologically inspired technique in which an excited (i.e., spiking) neuron reduces the relative spiking activity of its neighboring neurons. In this work, we apply 1-Winner Take All (1-WTA) lateral inhibition to the output spike times of the processing layer. The earliest spike is allowed to propagate whereas all other spikes are suppressed. The output spiketime $t_k^{(o)}$ generated after 1-WTA mechanism can be formally computed as follows:
\begin{align*}
    t_k^{(o)} &= \begin{cases}
    t_k & \text{if } t_k = t_{\min}, \;\nexists\;m < k \; \mathrm{s.t.}\; t_m = t_{\min}\\
    T_{\max} & \text{otherwise}
    \end{cases}
\end{align*}
Here, 
$t_{\min}$ is the earliest time at which any neuron in the processing layer fires. If multiple neurons spike at time $t_{\min}$, tie breaking selects the lowest index neuron. In case of no spike, the neuron with the maximum body potential is chosen.
\subsection{STDP Training}
\label{ssec:subsubhead}
A modified form of spike timing dependent plasticity (STDP) is used to train the synaptic weights of the TNN processing layer. In this work, we consider a stochastic, unsupervised learning rule with integer updates that is amenable to direct hardware implementation. 
The synaptic weight $w_{kj}$ linking neuron $j$ in the encoding layer to neuron $k$ in the processing layer is updated according to the following rules:
\begin{scriptsize}
\begin{table}[ht]
\centering
\scalebox{0.96}{
\begin{tabular}{|l|l|l|l|l|}
\hline
\multicolumn{2}{|c|}{\textbf{Spike Time Conditions}} & \multicolumn{3}{c|}{$\mathbf{\Delta w_{kj}}$} \\ \hline\hline
\multicolumn{2}{|l|}{$t_j^{(e)} \neq T_{\max},\; t_k^{(o)} = T_{\max}$} & \multicolumn{3}{l|}{$+X_s$} \\ \hline
\multicolumn{2}{|l|}{\begin{tabular}[c]{@{}l@{}}$t_j^{(e)} \neq T_{\max},\;t_k^{(o)} \neq T_{max},$\\$t_j^{(e)} \leq t_k^{(o)}$\end{tabular}} & \multicolumn{3}{l|}{$+X_c\cdot\max(S_P(w_{kj}), X_{\min})$} \\ \hline
\multicolumn{2}{|l|}{\begin{tabular}[c]{@{}l@{}}$t_j^{(e)} \neq T_{\max},\; t_k^{(o)} \neq T_{\max},$\\$t_j^{(e)} > t_k^{(o)}$\end{tabular}} & \multicolumn{3}{l|}{$-X_c\cdot\max(S_N(w_{kj}), X_{\min})$} \\ \hline
\multicolumn{2}{|l|}{\begin{tabular}[c]{@{}l@{}}$t_j^{(e)} = T_{\max},\;t_k^{(o)} \neq T_{\max}$\end{tabular}} & \multicolumn{3}{l|}{$-X_b\cdot\max(S_N(w_{kj}), X_{\min})$} \\ \hline
\multicolumn{2}{|l|}{$t_j^{(e)} = T_{\max},\; t_k^{(o)} = T_{\max}$} & \multicolumn{3}{l|}{$0$} \\ \hline
\end{tabular}
}
  \caption{STDP Update Rules \cite{nair2020direct}}
  \label{table:stdp}
\end{table}
\end{scriptsize}\\
Here, $X_s, X_c, X_b, X_{\min}$ are Bernoulli random variables with respective probabilities $\pi_s, \pi_c, \pi_b, \pi_{\min}$ of being one.
Intuitively if the output spike time of the encoding neuron is less than or equal to that of the processing neuron, the corresponding synaptic weight is increased due to positive correlation. If the encoding spiketime is greater, the synaptic weight is reduced since the spike from the encoding neuron could not have led to a spike in the output neuron. 
After each STDP update, the stochastic weights are clamped within the interval $[0, w_{\max}]$.  We restrict $\pi_s < \pi_c < \pi_b$ such that synaptic weights are conservatively increased. $S_P(\cdot)$ and $S_N(\cdot)$ are stabilizing random variables used to achieve a bimodal convergence of weights, and are characterized as:
\begin{align*}
    \mathbb{P}\left[S_P(w_{kj})=1\right] &= \left(\frac{w_{kj}}{w_{\max}}\right)\cdot\left(2 - \frac{w_{kj}}{w_{\max}}\right)\\
    \mathbb{P}\left[S_N(w_{kj}=1)\right] &= \left(1 - \frac{w_{kj}}{w_{\max}}\right)\cdot\left(1 + \frac{w_{kj}}{w_{\max}}\right)\\
\end{align*}

\section{Experimental Evaluation}
\label{sec:results}

\subsection{Methodology}
\label{ssec:experiment_setup}
We evaluate our approach on 36 time-series datasets from the UCR time series archive \cite{UCRArchive2018}, as done in \cite{Ma2019}, \cite{Zhang2018}. The datasets span a broad range of application domains including electrocardiogram (ECG) signals, human activity signals, image data converted to time series, food spectrographs and sensor recordings. 
The number of encoding neurons $E$ per time series feature is equal to 8. We choose $\ell = \lfloor L/8 \rfloor$, where $L$ is the signal length. Thus, the total number of inputs to the TNN is $E \cdot \ell \leq L$. The number of processing neurons is equal to the number of clusters. $T_{\max}$ and $w_{\max}$ are set to 16 and 7 respectively. \textit{We use a 3-bit integer resolution for weights.}

The TNN is trained over multiple epochs for each dataset until approximate weight convergence is achieved. The cluster assignment is taken to be the index of the spiking neuron after 1-WTA. The winning neuron's spike time suggests the ``confidence'' in the cluster prediction. \textit{Rand Index} ($\mathrm{RI}$) is used as in \cite{Ma2019} to measure clustering quality 
and is defined as:
\begin{align*}
    \mathrm{RI} = \frac{\alpha + \beta}{N(N-1)/2}
\end{align*}
$\alpha$ ($\beta$) is the number of data sample pairs with same (different) label(s) that are assigned to same (different) cluster(s). 
\begin{figure}[h]
    \centering
    \includegraphics[trim=0 0 0 7 0, clip, width = 225pt]{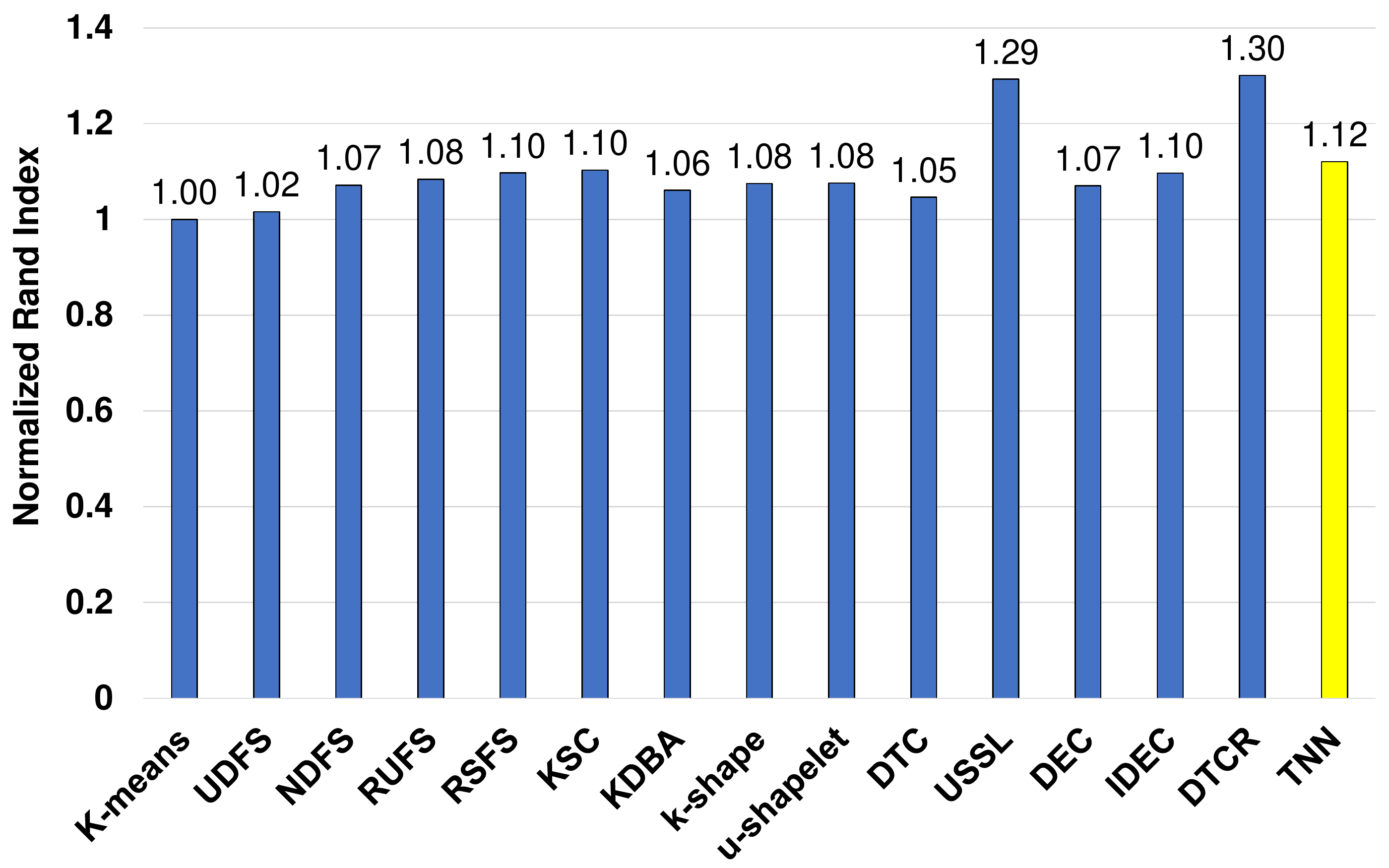}
    \caption{Comparison of TNN vs. state-of-the-art methods using Rand Index normalized to K-means}
    \label{fig:results_bar}
\end{figure}
\vspace{-15pt}
\subsection{Clustering Performance Results}
\label{ssec:results}
 The performance of TNN is compared to state-of-the-art algorithms reported in \cite{Ma2019} \cite{Zhang2018}. Figure \ref{fig:results_bar} depicts the rand index scores for all 15 algorithms (including TNN) averaged across the 36 datasets and normalized with respect to K-means. As shown in the figure, the proposed TNN approach outperforms 12 of the 14 algorithms even after the potential loss of information through random projections.
 The deep learning based approach DTCR \cite{Ma2019} and shapelet based approach USSL \cite{Zhang2018} still outperform the TNN.
 However, DTCR involves high-dimensional floating point tensor processing and  backpropagation. Meanwhile, USSL performs iterative coordinate descent in order to learn and refine shapelets according to pseudo labels. Both DTCR and USSL are far too computationally demanding for embedded edge devices.
 
  USSL is also not ideal for continuous streaming data as shapelets are learned over the entire training dataset. Changes in \textit{macro} patterns of the signal would require new shapelets to be explicitly learned. 
 In comparison, our approach uses only 3-bit integer weights, is extremely hardware efficient (next subsection) and capable of continuously learning and adapting to \textit{macro} level changes in the signal segments.
\begin{scriptsize}
\begin{table}[h]
  \centering
  \scalebox{1}{
  \begin{tabular}{|c|c|r|r|r|}
    \hline
    \textbf{TNN} & \textbf{Synapse} & \textbf{Area} & \textbf{Comp.} & \textbf{Power}\\
    \textbf{Design} & \textbf{Count} & [mm\textsuperscript{2}] & \textbf{Time} [ns] & [mW]\\
    \hline
    \hline
    Largest & 6750 & 0.033 & 6.50 & 0.155 \\
    \hline
    Smallest & 130 & 0.001 & 3.59 & 0.002 \\
    \hline
    \rowcolor{yellow} Average & 970 & 0.005 & 5.07 & 0.022 \\
    \hline
  \end{tabular}
  }
  \caption{Hardware Complexity for proposed TNN approach}
  \label{table:hw}
\end{table}
\end{scriptsize}
\vspace{-13pt}

\subsection{Hardware Complexity Analysis}
\label{sec:hardware}
We assess die area, power consumption and processing delay of the TNN processing layer, since the synapses in the TNN layer constitute majority of the hardware complexity. Table \ref{table:hw} provides the hardware complexity estimates for three TNN designs (average, largest and smallest) based on the characteristic equations and technology scaling from \cite{nair2020direct}. Our area and power metrics scale linearly with the synapse count while the latency scales logarithmically. Thus, the dimensionality reduction via random projections reduces the area and power by 77.5\% and the latency by up to 30\%. The average synapse count for the proposed TNN architectures across the 36 datasets is 970, which translates to about 5,000 $\mu$m\textsuperscript{2} die area, 22 $\mu$W power and 5 ns processing delay in 7 nm CMOS. The largest TNN network (\textit{WordSynonyms}) with 6,750 synapses incurs 33,000 $\mu$m\textsuperscript{2} and 155 $\mu$W. On the other hand, the smallest TNN network (\textit{SonyAIRobotSurface2}) with 130 synapses consumes less than 1,000 $\mu$m\textsuperscript{2} and 2 $\mu$W. 
\section{Conclusion}
\label{sec:conclusion}
We propose a TNN-based approach 
for online, unsupervised clustering of real-time sensory signals that is ideal for always-on, edge-native devices. The proposed architecture is capable of iterative continuous learning and can easily be scaled for multivariate signals. Our method outperforms K-means, and is competitive with  state-of-the-art algorithms when tested across a wide variety of time-series datasets. Furthermore, sizable TNNs can be directly implemented in 7 nm CMOS with less than 1 mm\textsuperscript{2} die area and 1 mW power. To the best of our knowledge, this is the first work that applies TNNs to the problem of univariate time-series clustering. This can serve as a foundation for future neuromorphic works in industrial edge applications such as anomaly detection and keyword spotting. 

\vfill\pagebreak

\bibliographystyle{IEEEbib}
\bibliography{final}

\end{document}